\documentclass{article}
\usepackage{spconf,amsmath,graphicx}
\usepackage{times}
\usepackage{soul}
\usepackage{url}
\usepackage[hidelinks]{hyperref}
\usepackage[utf8]{inputenc}
\usepackage[small]{caption}
\usepackage{graphicx}
\graphicspath{{picture/} }
\usepackage{amsmath}
\usepackage{multirow}
\usepackage{amsthm}
\usepackage{booktabs}
\usepackage{algorithm}
\usepackage{algorithmic}
\usepackage[switch]{lineno}
\usepackage{bbold}

\title{Contrastive Deep Nonnegative Matrix Factorization \\ for Community Detection}

\name{Yuecheng Li$^{1}$ \qquad Jialong Chen$^{1}$ \qquad Chuan Chen$^{1}$\sthanks{Corresponding author \\ This work was supported by the National Key Research and Development Program of China (2023YFB2703700), the National Natural Science Foundation of China (62176269, 12301411), the Natural Science Foundation of Guangdong (2023A1515012026), and the Guangzhou Science and Technology Program (2023A04J0314).} \qquad Lei Yang$^{1}$ \qquad Zibin Zheng$^{2}$ }
\address{$^{1}$ School of Computer Science and Engineering, Sun Yat-sen University, Guangzhou, China \\
$^{2}$ School of Software Engineering, Sun Yat-sen University, Zhuhai, China}

\begin{document}

%
\maketitle

\begin{abstract}
Recently, nonnegative matrix factorization (NMF) has been widely adopted for community detection, because of its better interpretability. However, the existing NMF-based methods have the following three problems: 1) they directly transform the original network into community membership space, so it is difficult for them to capture the hierarchical information; 2) they often only pay attention to the topology of the network and ignore its node attributes; 3) it is hard for them to learn the global structure information necessary for community detection. Therefore, we propose a new community detection algorithm, named \textbf{C}ontrastive \textbf{D}eep \textbf{N}onnegative \textbf{M}atrix \textbf{F}actorization (\textbf{CDNMF}). Firstly, we deepen NMF to strengthen its capacity for information extraction. Subsequently, inspired by contrastive learning, our algorithm creatively constructs network topology and node attributes as two contrasting views. Furthermore, we utilize a debiased negative sampling layer and learn node similarity at the community level, thereby enhancing the suitability of our model for community detection. We conduct experiments on three public real graph datasets and the proposed model has achieved better results than state-of-the-art methods. Code available at \url{https://github.com/6lyc/CDNMF.git}.
\end{abstract}
\section{Introduction}
\label{sec:intro}
Community detection (CD) is a fundamental task in complex network analysis. It involves partitioning a network into multiple substructures, each corresponding to a community. An effective partition requires nodes within the same community to be densely connected, while connections between nodes in different communities are sparse \cite{girvan2002community_first}. Mining the community structure is the key to revealing and understanding the organizational principles and operation of complex network systems. For example, in social networks, platforms detect different user communities to facilitate friend recommendation and advertisement placement \cite{xie2012socialnetwork}. In addition to applications in social networks, CD is also widely used in protein-protein interaction (PPI) networks \cite{gao2023PPI}, citation networks \cite{citation}, and more. Moreover, CD plays a crucial role in speaker diarization \cite{wang2023speaker1, zheng2022speaker2}.

Over the past two decades, many classical algorithms have been proposed for CD, such as modularity \cite{newman2006modularity}, conductance \cite{leskovec2010Conductance}, and permanence \cite{chakraborty2014permanence}. However, these methods are only capable of assigning each node to one community, thereby narrowing their scope of applicability \cite{ye2018danmf}. In recent years, certain neural network-based approaches have also been developed, including GUCD \cite{he2021GUCD} and VGAER \cite{qiu2022vgaer}. Nevertheless, their training processes resemble black boxes, leading to limited interpretability of the results. In addition, researchers have proposed community detection algorithms based on NMF \cite{he2021surveyNMF}, which have been widely used because of their good mathematical interpretability and natural applicability to the detection of overlapping communities. The NMF algorithm factorizes the adjacency matrix $A \in \mathbb{R}^{n\times n}_{+}$ of a graph in the following form: $A \approx U V $ s.t. $U \geq \mathbf{0}, V \geq \mathbf{0}$. Two nonnegative matrix factors $U \in \mathbb{R}_+^{n \times r}$ and $V \in \mathbb{R}_+^{r\times n}$ can be obtained. Pre-define the number of communities in the graph is $r$, the factor matrix $U$ can be regarded as a mapping between the original network and the community membership space, with each column $u_i (i=1,2,\cdots,r)$ of $U$ representing the basis vector of this space. The factor matrix $V$ can be regarded as the node representation matrix (community indication matrix), with its element $v_{i,j}$ quantifying the tendency of the $j$-th node to belong to the $i$-th community. Overall, the factorization process and results of the NMF algorithm demonstrate strong interpretability. Additionally, when performing overlapping CD, naturally, we can also assign each node to several communities with relatively high tendency based on the matrix $V$. There are also some variants of NMF with better performance, such as Orthogonal Nonnegative Matrix Factorization (ONMF) \cite{pompili2014ONMF}, Bayesian Nonnegative Matrix Factorization (BNMF) \cite{psorakis2011BNMF}, Nonnegative Matrix Tri-factorization (NMTF) \cite{pei2015NMTF} and MX-ONMTF \cite{ortiz2022MX-ONMTF} etc.

However, these methods still suffer from some of the following three problems:
\begin{enumerate}
    \item Such methods are shallow, often with only a single or double-layer mapping between the original network and the community membership space. Real-world networks have complex organizational principles, which are difficult to extract by shallow NMF.
    \item Such methods tend to focus only on the topology of the network while ignoring the attributes of the nodes. In social sciences, it has been shown that the node attributes can reflect and influence the structure of their communities \cite{mcpherson2001birds}.
    \item Such methods could not capture the community-level information in the network. NMF-based methods focus on the similarity between neighboring nodes and hardly extract global structural information necessary for CD.
  \end{enumerate}

This paper presents a new CD algorithm, called \textbf{CDNMF}, to address the limitations of existing methods. 
\begin{figure}[t]
    \centering
    \includegraphics[scale=0.33]{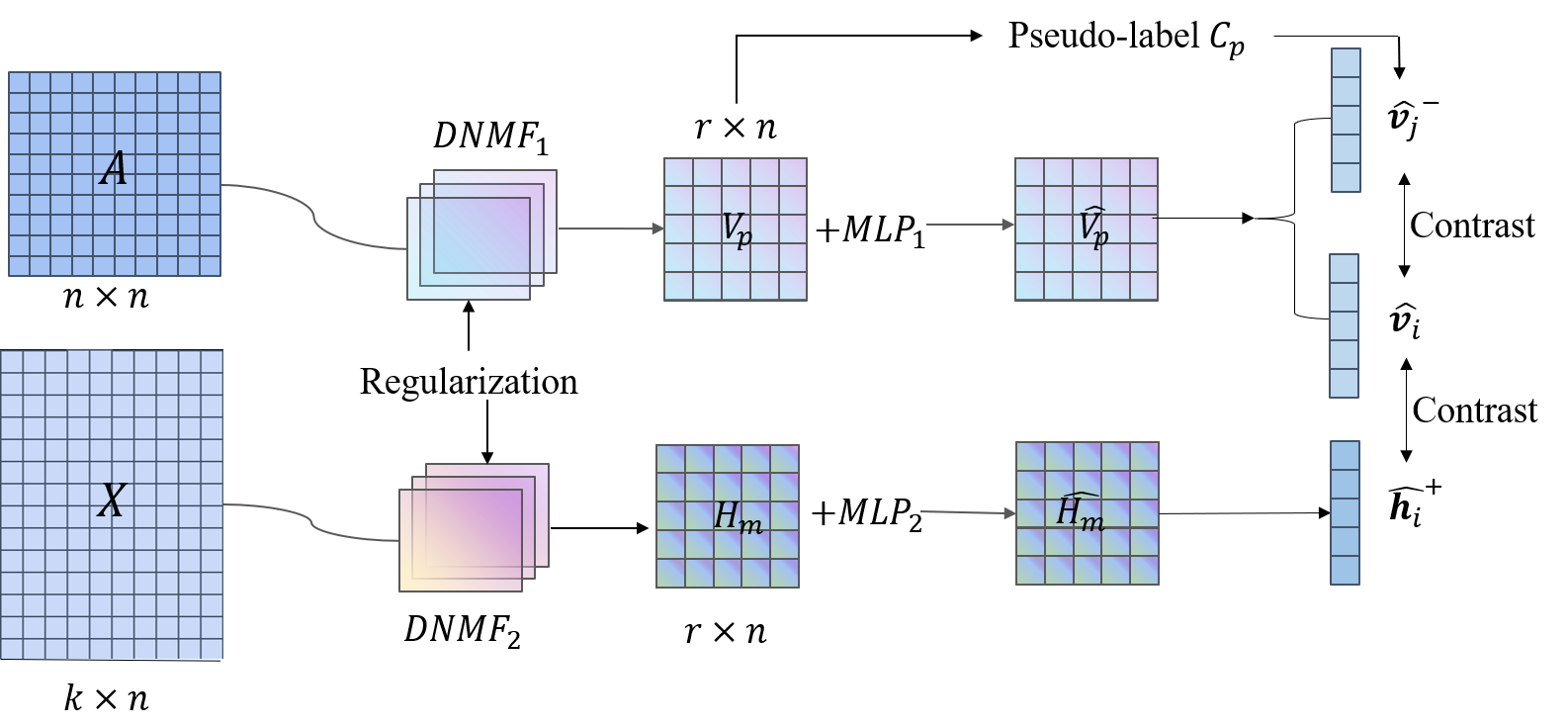}
    \caption{The general framework of our Contrastive Deep Nonnegative Matrix Factorization (CDNMF).}
    \label{g3}
\end{figure}
Overall, our main contributions are:
\begin{enumerate}
    \item We propose to use Deep NMF (DNMF) as the backbone to enhance the representation learning capability of our model.
    \item We construct a contrastive learning framework to unify the learning of graph topology and node attributes based on DNMF. Moreover, with the debiased negative sampling layer filtering out false negatives, our model better learns community-level information. 
    \item We conduct extensive experiments to evaluate the superiority, effectiveness, and efficiency of our CDNMF. The results show that our algorithm outperforms other SOTA CD methods.
\end{enumerate}

\section{Proposed Method}
\label{sec:method}
In this section, we describe our CDNMF in three modules. The general framework of the model is shown in Fig. \ref{g3}.

\subsection{DNMF Layer}
In order to extract the hierarchical information from the original network topology, we reconstruct the adjacency matrix $A$ as the deep form:
\begin{equation}
    \begin{aligned}
    & \min _{U_i, V_p} L_D=\left\|A-U_1 U_2 \ldots U_p V_p\right\|_F^2 \\
    & \text { s.t. } V_p \geq \mathbf{0}, U_i \geq \mathbf{0}, \forall i=1,2, \ldots, p, \label{eq1}
    \end{aligned}
\end{equation}
where $V_p \in \mathbb{R}_{+}^{r \times n}, U_i \in \mathbb{R}_{+}^{r_{i-1} \times r_i}(i=1,2, \ldots, p)$ and $n=r_0 \geq r_1 \geq \cdots \geq r_{p-1} \geq r_p=r$. $r$ is the number of communities in the network. The $\|\cdot\|_F$ denotes the Frobenius norm of the matrix. Each matrix $U_i(i=1,2, \ldots, p)$ can be interpreted as the $i$-th feature matrix with different hierarchical information. $U_1 U_2 \ldots U_n$ is the total mapping between the original network and the community membership space. The matrix $V_p$ is the node representation matrix after deep transformation, and each column can be understood as the tendency of a node to belong to different communities.

To transform the optimization problem with constraints in Eq. (\ref{eq1}) into an unconstrained problem, we design a penalty term for each nonnegative matrix. At first, we define the function $f: \mathbb{R}^{a \times b} \rightarrow \mathbb{R}^{a \times b}$ for any matrix $B \in \mathbb{R}^{a \times b}$ satisfying:
\begin{equation}
    f(B)=\left\{\begin{array}{c}
    B_{i j}, B_{i j}<0 \\
    \quad 0, B_{i j} \geq 0.
    \end{array}\right. \label{eq2}
\end{equation}
In brief, the function $f$ serves to convert the positive elements of the input matrix into 0 while leaving the negative elements unchanged. Naturally, the penalty term for the matrix $U_i(i=1,2, \ldots, p)$ is defined as:
\begin{equation}
    \min _{U_i}\left\|f\left(U_i\right)\right\|_F^2. \label{eq3}
\end{equation}
Similarly, the penalty term for the matrix $V_p$ is defined as:
\begin{equation}
    \min _{V_p}\left\|f\left(V_p\right)\right\|_F^2. \label{eq4}
\end{equation}

Combining Eq. (\ref{eq3}) as well as (\ref{eq4}), we transform the optimization problem Eq. (\ref{eq1}) into the following unconstrained objective function:
\begin{equation}
    \begin{aligned}
    & \min _{U_i, V_p} L_A=\left\|A-U_1 U_2 \ldots U_p V_p\right\|_F^2 \\
    & +\alpha\left(\sum_{i=1}^p\left\|f\left(U_i\right)\right\|_F^2+\left\|f\left(V_p\right)\right\|_F^2\right),
    \end{aligned} \label{eq5}
\end{equation}
where $\alpha > 0$ is the coefficient of the nonnegative penalty term. Similarly, for the node feature matrix $X$ of the network we obtain the following objective function:
\begin{equation}
    \begin{aligned}
    & \min _{W_j, H_m} L_X=\left\|X-W_1 W_2 \ldots W_m H_m\right\|_F^2 \\
    & +\alpha\left(\sum_{j=1}^m\left\|f\left(W_j\right)\right\|_F^2+\left\|f\left(H_m\right)\right\|_F^2\right). \label{eq6}
    \end{aligned}
\end{equation}

Overall, we obtain the objective function for the DNMF term:
\begin{equation}
    \min _{U_i, V_p, W_j, H_m} L_{D N M F} = L_A+L_X. \label{eq7}
\end{equation}

In addition, to preserve the intrinsic geometric structure of node pairs in deep hierarchical mapping, we introduce graph regularization as follows \cite{pei2015NMTF}:
\begin{equation}
    \frac{1}{2} \sum_i \sum_j A(i, j)\left\|V_p(:, i)-V_p(:, j)\right\|_2^2=\operatorname{tr}\left(V_p L V_p^T\right),
\end{equation}
where $L = D-A$ denotes the graph Laplacian matrix, $D$ is a diagonal matrix, and its diagonal elements are the row sums of $A$. $\operatorname{tr(\cdot)}$ denotes the trace of the matrix. 

Then, we minimize the graph regularization for two views and obtain the objective function for the graph regularization term:
\begin{equation}
    \begin{aligned}
    \min _{V_p, H_m} L_{r e g} & = L_{r e g_A}+L_{r e g_X} \\
    & =\operatorname{tr}\left(V_p L V_p^T\right)+\operatorname{tr}\left(H_m L H_m^T\right).
    \end{aligned}
\end{equation}

\subsection{Debiased Negative Sampling Layer}
After the DNMF layer, we would obtain the pseudo community labels from the node representation matrix, which could reduce the false negative samples. Specifically, we first obtain the pseudo labels of node $v_i$ by:
\begin{equation}
    c_i^*=\operatorname{argmax}\left(V_p(:, i)\right) \label{Eq10}
\end{equation}

Next, we remove all nodes that have the same pseudo label as $v_i$ to obtain the debiased negative sample set $\widetilde{\mathcal{N}}_i$ of node $v_i$:
\begin{equation}
    \widetilde{\mathcal{N}}_i=\left\{v_m\right\}\left(c_m^* \neq c_i^*\right). \label{eq11}
\end{equation}
In fact, the accuracy of pseudo community labels continues to improve with training iterations.

\subsection{Graph Contrastive Learning Layer}
As shown in Fig. \ref{g3}, we use the adjacency matrix $A$ of the network topology and the feature matrix $X$ of the node attributes as two views for contrastive learning. In particular, for each node $v_i$, we consider the representation vector $V_p(:, i)$ generated on the topology view as the anchor point, the representation vector $H_m(:, i)$ generated on the node attribute view as the positive sample, and the representation vectors of nodes in the set $\widetilde{\mathcal{N}}_i$ in Eq. (\ref{eq11}) as the negative samples. 

Then, for each positive sample pair $\left(V_p(:, i), H_m(:, i)\right)$, we define the contrastive loss as follows:
\begin{equation}
    \begin{aligned}
    & l\left(V_p(:, i), H_m(:, i)\right) \\
    & =\log \frac{e^{{\theta\left(V_p(:,i), H_m(:,i)\right)} / {\tau}}}{e^{{\theta\left(V_p(:,i),H_m(:,i)\right)} / {\tau}}+\sum_{k=1}^n \mathbb{1}_{\left[k \in \widetilde{\mathcal{N}_i}\right]} e^{{\theta\left(V_p(:,i),V_p(:,k)\right)} / {\tau}}}, \\
    &
    \end{aligned}
\end{equation}
where $\theta(\boldsymbol{v}, \boldsymbol{h})=s(g(\boldsymbol{v}), g(\boldsymbol{h}))$, $s$ is the cosine similarity function , and $g$ is the MLP of two layers. $\mathbb{1}_{\left[k \in \widetilde{\mathcal{N}}_i\right]} \in\{0,1\}$ is the indicator function that equals 1 if $k \in \widetilde{\mathcal{N}}_i$ and 0 otherwise. $\tau$ is the temperature parameter. 

The contrast of positive samples extracts consistency as well as complementary information from each node topology and attribute. The contrast of negative samples expands the distance between different communities so that our model learns the community-level similarity between nodes.

In the graph contrastive learning layer, we optimize the contrastive loss of each node, and obtain the objective function for the contrastive learning term:
\begin{equation}
    \begin{aligned}
    & \min _{V_p, H_m} L_{cl} =-\frac{1}{n} \sum_{i=1}^n l\left(V_p(:,i),H_m(:,i)\right)
    \end{aligned}
\end{equation}

\subsection{Training Process}
We jointly optimize each layer, and define the total objective function as follows:
\begin{equation}
    \min _{U_i, V_p, W_{j,}, H_m} L=L_{D N M F}+\beta L_{r e g}+\gamma L_{cl}, \label{eq14}
\end{equation}
where $\beta, \gamma>0$ are the scale factors. After the training process in Algorithm \ref{alg:algorithm}, the predicted community labels for each node $v_i$ are:
\begin{equation}
    C_p=\left\{c_i^*\right\}_{i=1}^n=\operatorname{argmax}\left(V_p\right).    
\end{equation}

\begin{algorithm}[tb]
    \caption{CDNMF}
    \label{alg:algorithm}
    \textbf{Input}: a graph $G=(A, X)$, the number of communities $r$;\\
    \textbf{Output}: $C_{p}$;
    \begin{algorithmic}[1] 
        \STATE \textbf{Pre-training stage}:
        \STATE $U_1, V_1 \leftarrow \operatorname{NMF}\left(A, r_1\right);$
        \STATE \textbf{for} $i=2$ to $p$ \textbf{do}
            \STATE \quad $U_i, V_i \leftarrow \operatorname{NMF}\left(V_{i-1}, r_i\right) ; / / r_p=r$
        \STATE \textbf{end for}
        \STATE The same pre-training process for $X$.
        \STATE \textbf{Fine-tuning stage}:
        \STATE Initialize each matrix factor with the pre-training values.
        \STATE \textbf{for} $i=1$ to $epoch$ \textbf{do} 
        \STATE \quad Generate the pseudo labels $c_i^*$ by Eq. (\ref{Eq10}).
        \STATE \quad Update the negative sample set $\widetilde{\mathcal{N}}_i$ by Eq. (\ref{eq11}).
        \STATE \quad Update model parameters by minimizing Eq.(\ref{eq14}) through SGD.
        \STATE \textbf{end for}
        \STATE \textbf{return} $C_{p}$
    \end{algorithmic}
\end{algorithm}

\section{Experiments}
\label{sec:pagestyle}
In this section, we first introduce our experimental setup and then compare our method with state-of-the-art methods in community detection tasks.
\subsection{Experimental Setup}
We perform experiments on three widely used graph datasets: Cora \cite{sen2008cora}, Citeseer \cite{rossi2015citeseer}, and PubMed \cite{namata2012pubmed}. Then, we compare our method with four types of CD methods, including four shallow NMF-based methods: NMF \cite{mankad2013NMF}, ONMF \cite{pompili2014ONMF}, BNMF \cite{psorakis2011BNMF}, NSED \cite{sun2017NSED}, 
three network embedding methods: LINE \cite{tang2015line}, Node2Vec \cite{grover2016node2vec}, MNMF \cite{wang2017ce}, 
three methods that consider node attributes: LP-FNMTF \cite{wang2011LP-FNMTF}, K-means++ \cite{vassilvitskii2006kmeans++}, VGAER \cite{qiu2022vgaer}
and two methods based on Deep NMF: DNMF \cite{ye2018danmf}, DANMF \cite{ye2018danmf}.

In particular, we fixed the number of hidden layers in the model to 3. The parameters used for the Cora are: $\alpha=400,\beta=3.0,\gamma=5.0,\tau=1.3$; the parameters used for the Citeseer are: $\alpha=3000,\beta=1.0,\gamma=5.0,\tau=1.5$; and the parameters used for the PubMed are: $\alpha=100,\beta=1.0,\gamma=1.0,\tau=0.5$. Moreover, ACC and NMI will be used for evaluating each CD method. We run each algorithm 20 times and report the average results.

\subsection{Community Detection}
It is worth noting that DNMF steadily obtains better performance than shallow NMF-based methods since it can learn hierarchical information. Our CDNMF obtains better performance than DNMF because it learns consistent semantic information from both network topology and node features. Moreover, our model expands the distance between different communities to fully learn the community-level similarity between nodes. Although all three embedding-based approaches attempt to maintain higher-order similarity between nodes, they do not exhibit competitive performance. For LINE and Node2Vec, they do not expressly model to preserve community-level similarity between nodes. In contrast, MNMF applies modularity to reveal the community structure of the network. However, the modularity approach may suffer from the Resolution Limit Problem \cite{fortunato2007resolution}. The results are shown in Table \ref{t1}.

\begin{table}[]
\centering
\caption{Community detection performance with ACC and NMI on three datasets. The \textbf{bold} and \underline{underlined} text indicate the optimal and suboptimal results, respectively.}
\label{t1}
\scalebox{0.78}{
\begin{tabular}{p{1.7cm}p{0.9cm}p{0.9cm}p{0.9cm}p{0.9cm}p{0.9cm}p{0.9cm}}
\hline
\multirow{2}{*}{Method} & \multicolumn{2}{c}{Cora}                                            & \multicolumn{2}{c}{Citeseer}                                        & \multicolumn{2}{c}{PubMed}                                          \\ \cline{2-7} 
                        & \,\,ACC                          & \,\,NMI                          & \,\,ACC                          & \,\,NMI                          & \,\,ACC                          & \,\,NMI                          \\ \hline
NMF                     & 0.4103                           & 0.2851                           & 0.3074                           & 0.1319                           & 0.5133                           & 0.1606                           \\
ONMF                    & 0.3811                           & 0.2416                           & 0.3330                           & 0.1423                           & 0.5575                           & 0.1582                           \\
BNMF                    & 0.4191                           & 0.2521                           & 0.3324                           & 0.0825                           & 0.5110                           & 0.0714                           \\
NSED                    & 0.4234                           & 0.2928                           & 0.3448                           & 0.1492                           & 0.5201                           & 0.1729                           \\ \hline
LINE                    & 0.4044                           & 0.2376                           & 0.3019                           & 0.0573                           & 0.4990                           & 0.1357                           \\
Node2Vec                & 0.3674                           & 0.1978                           & 0.2521                           & 0.0486                           & 0.4067                           & 0.0635                           \\
MNMF                    & 0.1647                           & 0.0035                           & 0.1890                           & 0.0031                           & 0.3397                           & 0.0002                           \\ \hline
LP-FNMTF                & 0.2861                           & 0.0261                           & 0.2327                           & 0.0143                           & 0.5437                           & 0.1532                           \\
K-means++               & 0.3230                           & 0.2210                           & 0.4160                           & 0.1910                           & 0.4150                           & \underline{0.2300}                           \\
VGAER               & 0.4530                           & 0.2970                           & 0.3020                           & \underline{0.2170}                           & 0.3010                           & 0.2230                           \\ 
\hline
DNMF                    & 0.4849                           & 0.3572                           & 0.3635                           & 0.1582                           & 0.5389                           & 0.1709                           \\
DANMF                   & \underline{0.5499}                           & \underline{0.3764}                           & \underline{0.4242}                           & 0.1831                           & \underline{0.6393}                           & 0.2221                           \\
Ours                    & \textbf{0.6081} & \textbf{0.4006} & \textbf{0.4756} & \textbf{0.2559} & \textbf{0.6653} & \textbf{0.2330} \\ \hline
\end{tabular}}
\end{table}

\begin{table}[]
\centering
\caption{Results of ablation experiments based on Cora and Citeseer.}
\label{t2}
\centering
\scalebox{0.78}{
\begin{tabular}{l p{0.6cm} p{0.6cm} p{0.6cm} p{0.6cm} p{0.6cm} p{0.6cm} p{0.6cm} p{0.6cm}}
\hline
\multirow{2}{*}{Methods}   & \multicolumn{4}{c}{Cora} & \multicolumn{4}{c}{Citeseer} \\ \cline{2-9} 
                           & \,\,ACC   &\,\,$\Delta$      & \,\,NMI  &\,\,$\Delta$      & \,\,ACC   &\,\,$\Delta$        & \,\,NMI    &\,\,$\Delta$      \\ \hline
Ours $[L(A)]$ & 0.5835      & \small2.46\% & 0.3781  & \small2.25\%   & 0.4598 & \small1.58\%       & 0.1672  & \small8.87\%     \\
Ours $[L(X)]$ & 0.5162  & \small9.19\%    & 0.3501 & \small5.05\%    & 0.3499  & \small12.6\%      & 0.1749  & \small8.10\%     \\ \hline
Ours                       & \textbf{0.6081}  &     & \textbf{0.4006} &     & \textbf{0.4756}   &      & \textbf{0.2559} &       \\ \hline
\end{tabular}}
\end{table}

\subsection{Ablation Experiments}
We consider removing our model's graph contrastive learning layer, i.e., let $\gamma$ = 0 in the total objective function. Then we perform community detection experiments based on the adjacency matrix $A$ and the node feature matrix $X$, respectively, denoted as "Ours $[L(A)]$" and "Ours $[L(X)]$". 

In both datasets, "Ours $[L(A)]$" basically outperforms "Ours $[L(X)]$", which indicates that the topology of the network has more influence on community detection to some extent. However, both of them are less effective than "Ours". This indicates that the information from a single view, either the topology of the network or the attributes of the nodes, cannot accurately model the community-level relationships between nodes. The results are shown in Table \ref{t2}.

\subsection{Convergence Analysis }
\begin{figure}[t]
    \centering
    \scalebox{0.9}{\includegraphics[width=3.5in]{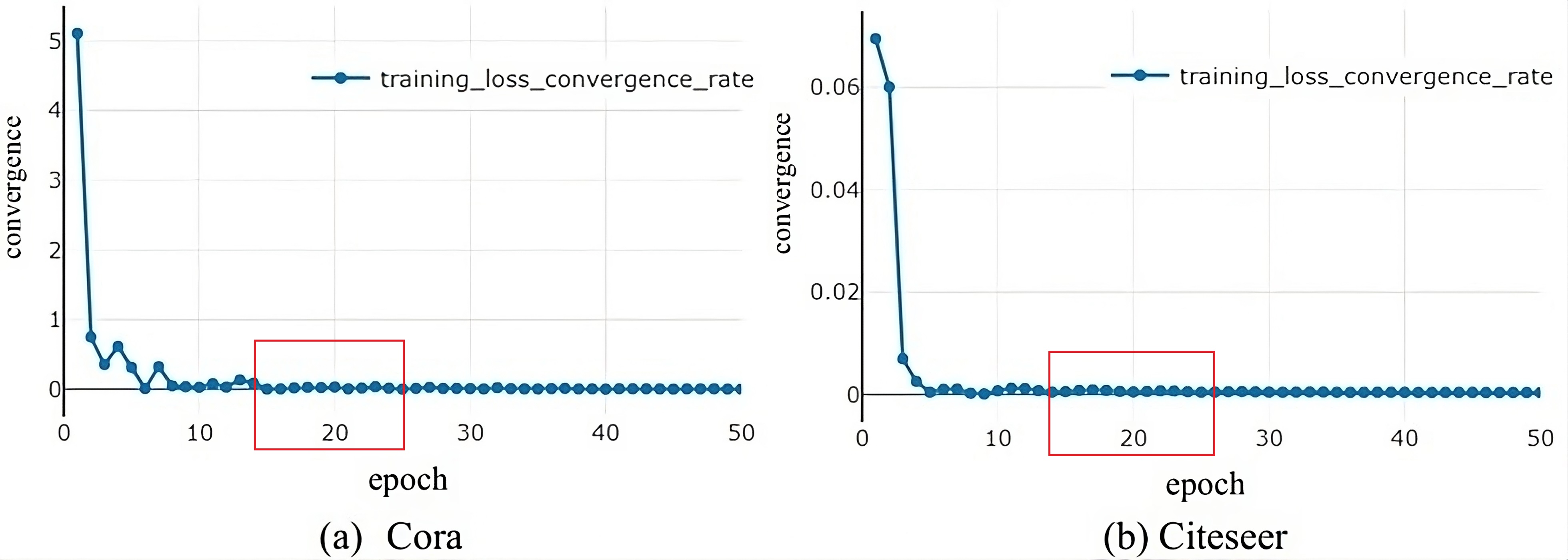}}
    \caption{The analysis of the convergence rate of our algorithm.}
    \label{g6}
\end{figure}

 Our optimization process is divided into two stages. In the pre-training stage, we perform NMF for each layer of the two input matrices, and its convergence has been analyzed in \cite{sun2017NSED}. Next, we visualize the convergence of the fine-tuning stage in Fig. \ref{g6}. We observe that our model can achieve convergence within about 20 epochs with a fast convergence rate. Similar results also can be observed on PubMed.      

\section{Conclusion}
In this paper, we propose a novel community detection method called CDNMF, which innovatively combines the idea of graph contrastive learning into NMF. Our CDNMF extracts the consistency information in network topology and node features through positive sample pairs and expands the distance of different communities in the representation space through negative sample pairs. We conduct various experiments to verify the superiority, effectiveness, and efficiency of our model. Moreover, there is potential to integrate additional matrix factorization and contrastive learning algorithms.

\vfill\pagebreak
\footnotesize
\bibliographystyle{IEEEbib}
\bibliography{paper}

\begin{thebibliography}{10}

\bibitem{girvan2002community_first}
Michelle Girvan and Mark~EJ Newman,
\newblock ``Community structure in social and biological networks,''
\newblock {\em Proceedings of the national academy of sciences}, vol. 99, no. 12, pp. 7821--7826, 2002.

\bibitem{xie2012socialnetwork}
Jierui Xie and Boleslaw~K Szymanski,
\newblock ``Towards linear time overlapping community detection in social networks,''
\newblock in {\em Pacific-Asia Conference on Knowledge Discovery and Data Mining}. Springer, 2012, pp. 25--36.

\bibitem{gao2023PPI}
Ziqi Gao, Chenran Jiang, Jiawen Zhang, Xiaosen Jiang, Lanqing Li, Peilin Zhao, Huanming Yang, Yong Huang, and Jia Li,
\newblock ``Hierarchical graph learning for protein--protein interaction,''
\newblock {\em Nature Communications}, vol. 14, no. 1, pp. 1093, 2023.

\bibitem{citation}
Fanzhen Liu, Shan Xue, Jia Wu, Chuan Zhou, Wenbin Hu, Cecile Paris, Surya Nepal, Jian Yang, and Philip~S Yu,
\newblock ``Deep learning for community detection: Progress, challenges and opportunities,''
\newblock in {\em Proceedings of the Twenty-Ninth International Joint Conference on Artificial Intelligence}, July 2020, IJCAI-PRICAI-2020.

\bibitem{wang2023speaker1}
Jie Wang, Zhicong Chen, Haodong Zhou, Lin Li, and Qingyang Hong,
\newblock ``Community detection graph convolutional network for overlap-aware speaker diarization,''
\newblock in {\em ICASSP 2023-2023 IEEE International Conference on Acoustics, Speech and Signal Processing (ICASSP)}. IEEE, 2023, pp. 1--5.

\bibitem{zheng2022speaker2}
Siqi Zheng and Hongbin Suo,
\newblock ``Reformulating speaker diarization as community detection with emphasis on topological structure,''
\newblock in {\em ICASSP 2022-2022 IEEE International Conference on Acoustics, Speech and Signal Processing (ICASSP)}. IEEE, 2022, pp. 8097--8101.

\bibitem{newman2006modularity}
Mark~EJ Newman,
\newblock ``Modularity and community structure in networks,''
\newblock {\em Proceedings of the national academy of sciences}, vol. 103, no. 23, pp. 8577--8582, 2006.

\bibitem{leskovec2010Conductance}
Jure Leskovec, Kevin~J Lang, and Michael Mahoney,
\newblock ``Empirical comparison of algorithms for network community detection,''
\newblock in {\em Proceedings of the 19th international conference on World wide web}, 2010, pp. 631--640.

\bibitem{chakraborty2014permanence}
Tanmoy Chakraborty, Sriram Srinivasan, Niloy Ganguly, Animesh Mukherjee, and Sanjukta Bhowmick,
\newblock ``On the permanence of vertices in network communities,''
\newblock in {\em Proceedings of the 20th ACM SIGKDD international conference on Knowledge discovery and data mining}, 2014, pp. 1396--1405.

\bibitem{ye2018danmf}
Fanghua Ye, Chuan Chen, and Zibin Zheng,
\newblock ``Deep autoencoder-like nonnegative matrix factorization for community detection,''
\newblock in {\em Proceedings of the 27th ACM international conference on information and knowledge management}, 2018, pp. 1393--1402.

\bibitem{he2021GUCD}
Dongxiao He, Yue Song, Di~Jin, Zhiyong Feng, Binbin Zhang, Zhizhi Yu, and Weixiong Zhang,
\newblock ``Community-centric graph convolutional network for unsupervised community detection,''
\newblock in {\em Proceedings of the twenty-ninth international conference on international joint conferences on artificial intelligence}, 2021, pp. 3515--3521.

\bibitem{qiu2022vgaer}
Chenyang Qiu, Zhaoci Huang, Wenzhe Xu, and Huijia Li,
\newblock ``Vgaer: graph neural network reconstruction based community detection,''
\newblock {\em arXiv preprint arXiv:2201.04066}, 2022.

\bibitem{he2021surveyNMF}
Chaobo He, Xiang Fei, Qiwei Cheng, Hanchao Li, Zeng Hu, and Yong Tang,
\newblock ``A survey of community detection in complex networks using nonnegative matrix factorization,''
\newblock {\em IEEE Transactions on Computational Social Systems}, 2021.

\bibitem{pompili2014ONMF}
Filippo Pompili, Nicolas Gillis, P-A Absil, and Fran{\c{c}}ois Glineur,
\newblock ``Two algorithms for orthogonal nonnegative matrix factorization with application to clustering,''
\newblock {\em Neurocomputing}, vol. 141, pp. 15--25, 2014.

\bibitem{psorakis2011BNMF}
Ioannis Psorakis, Stephen Roberts, Mark Ebden, and Ben Sheldon,
\newblock ``Overlapping community detection using bayesian non-negative matrix factorization,''
\newblock {\em Physical Review E}, vol. 83, no. 6, pp. 066114, 2011.

\bibitem{pei2015NMTF}
Yulong Pei, Nilanjan Chakraborty, and Katia Sycara,
\newblock ``Nonnegative matrix tri-factorization with graph regularization for community detection in social networks,''
\newblock in {\em Twenty-fourth international joint conference on artificial intelligence}, 2015.

\bibitem{ortiz2022MX-ONMTF}
Meiby Ortiz-Bouza and Selin Aviyente,
\newblock ``Orthogonal nonnegative matrix tri-factorization for community detection in multiplex networks,''
\newblock in {\em ICASSP 2022-2022 IEEE International Conference on Acoustics, Speech and Signal Processing (ICASSP)}. IEEE, 2022, pp. 5987--5991.

\bibitem{mcpherson2001birds}
Miller McPherson, Lynn Smith-Lovin, and James~M Cook,
\newblock ``Birds of a feather: Homophily in social networks,''
\newblock {\em Annual review of sociology}, pp. 415--444, 2001.

\bibitem{sen2008cora}
Prithviraj Sen, Galileo Namata, Mustafa Bilgic, Lise Getoor, Brian Galligher, and Tina Eliassi-Rad,
\newblock ``Collective classification in network data,''
\newblock {\em AI magazine}, vol. 29, no. 3, pp. 93--93, 2008.

\bibitem{rossi2015citeseer}
Ryan Rossi and Nesreen Ahmed,
\newblock ``The network data repository with interactive graph analytics and visualization,''
\newblock in {\em Proceedings of the AAAI conference on artificial intelligence}, 2015, vol.~29.

\bibitem{namata2012pubmed}
Galileo Namata, Ben London, Lise Getoor, Bert Huang, and U~Edu,
\newblock ``Query-driven active surveying for collective classification,''
\newblock in {\em 10th international workshop on mining and learning with graphs}, 2012, vol.~8, p.~1.

\bibitem{mankad2013NMF}
Shawn Mankad and George Michailidis,
\newblock ``Structural and functional discovery in dynamic networks with non-negative matrix factorization,''
\newblock {\em Physical Review E}, vol. 88, no. 4, pp. 042812, 2013.

\bibitem{sun2017NSED}
Bing-Jie Sun, Huawei Shen, Jinhua Gao, Wentao Ouyang, and Xueqi Cheng,
\newblock ``A non-negative symmetric encoder-decoder approach for community detection,''
\newblock in {\em Proceedings of the 2017 ACM on Conference on Information and Knowledge Management}, 2017, pp. 597--606.

\bibitem{tang2015line}
Jian Tang, Meng Qu, Mingzhe Wang, Ming Zhang, Jun Yan, and Qiaozhu Mei,
\newblock ``Line: Large-scale information network embedding,''
\newblock in {\em Proceedings of the 24th international conference on world wide web}, 2015, pp. 1067--1077.

\bibitem{grover2016node2vec}
Aditya Grover and Jure Leskovec,
\newblock ``node2vec: Scalable feature learning for networks,''
\newblock in {\em Proceedings of the 22nd ACM SIGKDD international conference on Knowledge discovery and data mining}, 2016, pp. 855--864.

\bibitem{wang2017ce}
Xiao Wang, Peng Cui, Jing Wang, Jian Pei, Wenwu Zhu, and Shiqiang Yang,
\newblock ``Community preserving network embedding,''
\newblock in {\em Thirty-first AAAI conference on artificial intelligence}, 2017.

\bibitem{wang2011LP-FNMTF}
Hua Wang, Feiping Nie, Heng Huang, and Fillia Makedon,
\newblock ``Fast nonnegative matrix tri-factorization for large-scale data co-clustering,''
\newblock in {\em Twenty-Second International Joint Conference on Artificial Intelligence}, 2011.

\bibitem{vassilvitskii2006kmeans++}
Sergei Vassilvitskii and David Arthur,
\newblock ``k-means++: The advantages of careful seeding,''
\newblock in {\em Proceedings of the eighteenth annual ACM-SIAM symposium on Discrete algorithms}, 2006, pp. 1027--1035.

\bibitem{fortunato2007resolution}
Santo Fortunato and Marc Barth{\'e}lemy,
\newblock ``Resolution limit in community detection,''
\newblock {\em Proceedings of the National Academy of Sciences of the United States of America}, pp. 36--41, 2007.

\end{thebibliography}

\end{document}